\pdfoutput=1

\documentclass[11pt]{article}

\usepackage[final]{acl}

\usepackage{times}
\usepackage{latexsym}

\usepackage[T1]{fontenc}

\usepackage[utf8]{inputenc}

\usepackage{microtype}

\usepackage{inconsolata}

\usepackage{graphicx}

\usepackage{adjustbox}
\usepackage{makecell}
\usepackage{multirow}
\usepackage[ruled,vlined,linesnumbered]{algorithm2e}
\usepackage{enumitem}
\usepackage{subcaption}
\usepackage{amsmath,amsthm,amsfonts}
\usepackage{xcolor}
\definecolor{oceanboatblue}{rgb}{0.0, 0.47, 0.75}
\usepackage{xspace}
\usepackage{tcolorbox}
\usepackage{booktabs}
\usepackage{lipsum} 

\usepackage{tikz}
\usepackage{pgfplots,pgfplotstable}
\usepgfplotslibrary[groupplots]
\usetikzlibrary{patterns, shapes.geometric, arrows, arrows.meta}

\usepackage{mdframed}
\newmdtheoremenv[%
  backgroundcolor=gray!20,
  linecolor=red!60!black,
  linewidth=2pt,
  topline=false,
  rightline=false,
  skipabove=10pt,
  skipbelow=10pt,
  leftline=false]{regbox}{Box}

\newcommand{\specialcell}[2][c]{%
    \begin{tabular}[#1]{@{}c@{}}#2\end{tabular}
}
\newenvironment{mybox}{
\definecolor{lightgray}{RGB}{220,220,220}
\begin{tcolorbox}[colback=lightgray,colframe=black,arc=4pt]
}{
\end{tcolorbox}
}

\definecolor{cycle1}{RGB}{228, 26, 28}
\definecolor{cycle2}{RGB}{55, 126, 184}
\definecolor{cycle3}{RGB}{77, 175, 74}
\definecolor{cycle4}{RGB}{152, 78, 163}
\definecolor{cycle5}{RGB}{0, 32, 96}
\definecolor{cycle6}{RGB}{153, 153, 153}
\definecolor{cycle7}{RGB}{166, 86, 40}
\definecolor{cycle8}{RGB}{247, 129, 191}
\definecolor{cycle9}{RGB}{252, 217, 60}
\definecolor{cycle10}{RGB}{27, 29, 191}
\definecolor{cycle11}{RGB}{255, 0, 0}

\newcommand{\eg}{\emph{e.g.}}
\newcommand{\ie}{\emph{i.e.}}
\newcommand{\etc}{\emph{etc}}
\newcommand{\olddataset}{\textsc{ChEBI-20}\xspace}
\newcommand{\oldmodel}{\textsc{MolT5}\xspace}
\newcommand{\newdataset}{\textsc{LaChEBI-20}\xspace}
\newcommand{\newmodel}{\textsc{LaMolT5}\xspace}
\newcommand{\pipeline}{\textsc{LA$^3$}\xspace}
\newcommand{\bace}{\texttt{ogbg-molbace}\xspace}
\newcommand{\hiv}{\texttt{ogbg-molhiv}\xspace}
\newcommand{\esol}{\texttt{ogbg-molesol}\xspace}
\newcommand{\image}{\texttt{CC3M}\xspace}
\newcommand{\spara}[1]{\smallskip\noindent{\bf #1}}

\newcommand{\para}[1]{\noindent{\bf #1}}

\title{
Automatic Annotation Augmentation\\Boosts Translation between Molecules and Natural Language
}

\author{
 \textbf{Zhiqiang Zhong\textsuperscript{1}*},
 \textbf{Simon Sataa-Yu Larsen\textsuperscript{1}*},
 \textbf{Haoyu Guo\textsuperscript{1}*},
 \textbf{Tao Tang\textsuperscript{1}*},
\\
 \textbf{Kuangyu Zhou\textsuperscript{2}},
 \textbf{Davide Mottin\textsuperscript{1}},
\\
\\
 \textsuperscript{1}Aarhus University,
 \textsuperscript{2}Microsoft,
\\
   \textbf{Correspondence:} \href{mailto:zzhong@cs.au.dk}{zzhong@cs.au.dk}
   \;
   *: Equal Contribution 
}

\begin{document}
\maketitle

\begin{abstract} 
Recent advancements in AI for biological research focus on integrating molecular data with natural language to accelerate drug discovery. 
However, the scarcity of high-quality annotations limits progress in this area. 
This paper introduces \pipeline, a Language-based Automatic Annotation Augmentation framework that leverages large language models to augment existing datasets, thereby improving AI training. 
We demonstrate the effectiveness of \pipeline by creating an enhanced dataset, \newdataset, where we systematically rewrite the annotations of molecules from an established dataset. 
These rewritten annotations preserve essential molecular information while providing more varied sentence structures and vocabulary. 
Using \newdataset, we train \newmodel based on a benchmark architecture to learn the mapping between molecular representations and augmented annotations.

Experimental results on text-based \emph{de novo} molecule generation and molecule captioning demonstrate that \newmodel outperforms state-of-the-art models. Notably, incorporating \pipeline leads to improvements of up to $301\%$ over the benchmark architecture.
Furthermore, we validate the effectiveness of \pipeline notable applications in \emph{image}, \emph{text} and \emph{graph} tasks, affirming its versatility and utility.
\footnote{The augmented dataset and trained models are available at \url{https://github.com/zhiqiangzhongddu/LA3}.}

\end{abstract}

\section{Introduction} 
\label{sec:introduction}
\begin{figure}[!ht]
    \centering
    \begin{tikzpicture}
        \begin{axis}[
            width=7.5cm, height=6cm,
            xlabel={Epoch},
            ylabel={BLEU},
            xmin=100, xmax=1000,
            ymin=0.5, ymax=0.87,
            grid=major,
            legend cell align={left},
            legend columns=2,
            legend style={font=\scriptsize,at={(0.6,-0.0)},anchor=south, font=\scriptsize, draw=none},
            xlabel near ticks,
            ylabel near ticks,
            tick label style={font=\scriptsize},
            label style={font=\small},
        ]
            \draw[>=triangle 45, <->, black] (880,255) -- (880,350) node[below, midway, xshift=-0.5cm, yshift=-0.65cm, purple] {\small \textbf{+12.3\%}};
            
            \addplot+[
                purple, mark=diamond*, mark options={solid}
            ] coordinates {
                (100,0.711) (200,0.725) (300,0.779) (400,0.811) (500,0.820) (600,0.831) (700,0.837) (800,0.841) (900,0.846) (1000,0.848)
            };
            \addlegendentry{\pipeline}
            
            \addplot+[
                teal, mark=square, mark options={solid}
            ] coordinates {
                (100,0.729) (200,0.734) (300,0.732) (400,0.736) (500,0.737) (600,0.748) (700,0.759) (800,0.769) (900,0.770) (1000,0.771)
            };
            \addlegendentry{EDA+Mixup}
            
            \addplot+[
                olive, mark=*, mark options={solid}
            ] coordinates {
                (100,0.691) (200,0.704) (300,0.732) (400,0.766) (500,0.771) (600,0.784) (700,0.799) (800,0.807) (900,0.817) (1000,0.817)
            };
            \addlegendentry{\pipeline: GPT}

            \addplot+[
                orange, mark=star, mark options={solid}
            ] coordinates {
                (100,0.611) (200,0.644) (300,0.671) (400,0.675) (500,0.676) (600,0.657) (700,0.661) (800,0.660) (900,0.661) (1000,0.661)
            };
            \addlegendentry{Generation}
            
            \addplot+[
                blue, mark=triangle*, mark options={solid}
            ] coordinates {
                (100,0.660) (200,0.694) (300,0.710) (400,0.745) (500,0.761) (600,0.767) (700,0.771) (800,0.798) (900,0.799) (1000,0.798)
            };
            \addlegendentry{\pipeline: Gemini}
            
            \addplot[black, thick, dashed] coordinates {(100,0.755) (1000,0.755)} node[pos=0.1, above] {\scriptsize \oldmodel};
            
        \end{axis}
    \end{tikzpicture}
    \caption{
    Molecule generation performance of \newmodel-Small with different \pipeline augmentations.
    Conventional augmentation (EDA~\cite{WZ19}, Mixup~\cite{ZCDL18}) and straightforward LLMs for data generation~\cite{ZZM24} fall behind. 
    }
    \label{fig:ablation}
\vspace{-3mm}
\end{figure}
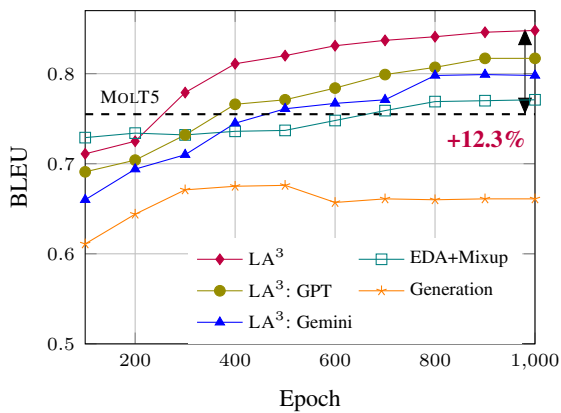

Artificial Intelligence (AI) has garnered increasing attention due to its transformative potential in broad real-world applications, including biology~\cite{SLGBHBL19,PZZWGWXY23,ZM23}. 
Take a recent new trend as an example: researchers intend to jointly model SMILES~\cite{W88}
strings and scientific text to obtain shared representations across the two modalities.
For instance, \citet{ELRHCJ22} innovatively propose \oldmodel, a model based on the T5 architecture~\cite{RSRLNMZLL20}, pre-trained on ZINC~\cite{SI15} by predicting the masked text parts.  
Consequently, they fine-tune the model on \olddataset~\cite{EZJ21} to learn how to map between SMILES representations of molecules and their corresponding annotations (captions) to support molecular tasks. 

However, the development of AI faces a fundamental setback: \emph{the scarcity of high-quality annotated data}. 
For instance, molecular data annotation is often a costly and time-consuming process~\cite{DGH16}. 
This limitation restrains the development of AI approaches, as models grow in size and expressiveness, they require larger and more diverse annotated datasets to achieve high performance and generalisability~\cite{DCLT19,HGC23}.
Therefore, one viable alternative for AI in practice is to resort to \emph{effective data augmentation strategies.}

Back to the example \olddataset dataset, a set of studies attempted various solutions to address the scarcity limitation, following the \oldmodel. 
\citet{CGBWLM23}, \citet{LZXWXQZL23}, \citet{PZZWGWXY23} and \citet{PWGLFZXQY24} introduce extra chemical databases and auxiliary tasks to train advanced models. 
However, the heavy dependencies on supplemental datasets limit their practical versatility, and the auxiliary task definition requires domain expertise. 
On the other hand, \citet{LLFWLTL24} use the general human knowledge embedded in Large Language Models (LLMs) to perform the molecule-caption translation tasks. 
Nevertheless, despite the widespread use of LLMs in literature review, table interpretation, \etc, their application in biology is not straightforward \cite{LJRHHNPWR24}.
The effectiveness of \cite{LLFWLTL24} depends heavily on the specific retrieval strategy used, and its performance is surpassed by existing smaller models. 
Additionally, \citet{ZZM24} and our empirical studies in Figures~\ref{fig:ablation},\ref{fig:compare_sota} demonstrate the limitations of using LLMs for data augmentation.

This paper proposes an effective automatic pipeline (see Figure~\ref{fig:rewrite_training_pipeline}), \pipeline, to effectively augment annotations of datasets with no human supervision.
Once the augmented datasets are generated, existing methods can be conveniently re-trained for significant performance boosting. 

We showcase the effectiveness of \pipeline by creating an enhanced dataset, \newdataset, where we leverage the in-context learning~\cite{LYFJ23} capability of LLMs to rewrite the annotation of each molecule in \olddataset. 
These rewritten annotations preserve essential molecular information while providing more varied sentence structures and vocabulary (see our analysis in Section~\ref{subsec:analysis}). 
After the annotation augmentation process, each molecule in \newdataset is accompanied by diverse annotations. 
Using these annotations, we proceed to train \newmodel, using the benchmark \oldmodel architecture, to support molecular tasks. 
During training, \newmodel aims to learn a mapping function between the space of molecules and the augmented annotations, thereby enhancing the overall performance of the models.

We systematically evaluate the effectiveness of \newmodel on challenging text-based \emph{de novo} molecule generation and molecule captioning tasks. 
Through extensive experiments on the benchmark evaluation pipeline, we demonstrate that \newmodel significantly elevates the performance of \oldmodel, which was trained using the same architecture on the original \olddataset dataset. 
Notably, \newmodel achieves improvements of up to $301\%$ on the molecule generation task and $9.51\%$ on the molecule captioning task.
Additionally, the small-size variant of \newmodel ($77M$ parameters) outperforms the large-size variant of \oldmodel ($800M$ parameters) for the molecule generation task. 
Compared with other leading methods reported on the leaderboard, \newmodel achieves new state-of-the-art performance with $99\%$ fewer parameters.
More importantly, \pipeline effectively boosts the performance of other applications, including \emph{image captioning}, \emph{text understanding} and \emph{graph property prediction}, affirming its versatility.

\smallskip 
\noindent \textbf{Our contributions} are as follows: 
\textbf{(1)} A fully automated pipeline for domain-specific applications where limited data availability restricts the effectiveness of existing technologies. 
\textbf{(2)} A set of lightweight open-source models tailored to address challenging molecular tasks. 
\textbf{(3)} Empirical studies demonstrating the necessity and effectiveness of \pipeline across multiple applications.


\section{Related Work} 
\label{sec:related_work}
\spara{Molecule Language Models.} 
MLMs have recently seen significant advancements, leveraging NLP techniques to understand and generate molecules. 
Early works such as ChemBERTa~\cite{CGR20} and Text2Mol~\cite{EZJ21} adapt transformer-based architectures for molecular representation learning. 
MolGPT~\cite{BAVP22} and \oldmodel~\cite{ELRHCJ22} demonstrate the ability to predict molecular properties and generate novel compounds, highlighting the potential of language models in biological research. 
However, the effectiveness of MLMs is often constrained by the limited availability of annotated molecular data.
Meanwhile, manual molecular data annotation is often a costly and time-consuming process, necessitating specialised equipment and extensive human labour~\cite{DGH16}. 

\spara{Data Augmentation in MLMs.}
Data augmentation has emerged as a critical strategy to address the scarcity of high-quality datasets.
Take the following studies on \olddataset, \citet{CGBWLM23}, \citet{LZXWXQZL23}, \citet{PZZWGWXY23} and \citet{PWGLFZXQY24} introduce additional chemical databases (PubChem~\cite{KCCGHHLSTY23}, Drugbank~\cite{WFGLMGSJLS18}, UniProt~\cite{U23}, PubMed~\cite{W20}, \etc) as to enrich model with extra knowledge, and design auxiliary tasks to train advanced models. 
However, these methods depend on supplemental datasets and the domain expertise required to shape the tasks.
On the other hand, LLMs have experienced exponential growth in both size and capability in recent years~\cite{BMRS20}. 
A wide range of NLP applications have been reshaped by LLMs~\cite{AMAV23,ZZM242}.
Notably, MolReGPT~\cite{LLFWLTL24} leverage the built-in general human knowledge of LLMs to perform the molecule-caption translation tasks. 
Despite the widespread use of LLMs for tasks like literature review and table interpretation \cite{AAAA23,TMSA23}, their application in biology remains challenging \cite{LJRHHNPWR24}. The effectiveness of MolReGPT is highly dependent on the chosen retrieval strategy, and it is often outperformed by smaller, existing models.
 
In another related work, \citet{WYHYMW24} utilise LLMs and human annotators to augment text for each image to improve contrastive learning of CLIP~\cite{RKHRGASAMCKS21}, yet these pipelines still require human supervision.
Furthermore, \citet{ZZM24} reveal the limited capability of LLMs in understanding domain-specific data, \eg, biology and physics, making LLM-based synthetic data generation challenging in many applications (see our analysis in Section~\ref{subsec:analysis}). 
In contrast, our novel \pipeline pipeline is fully automated and tailored for domain-specific applications where data scarcity limits the effectiveness of current technologies.


\section{Methodology}
\label{sec:methodology}
\begin{figure}[!ht]
\centering
\includegraphics[width=.5\textwidth]{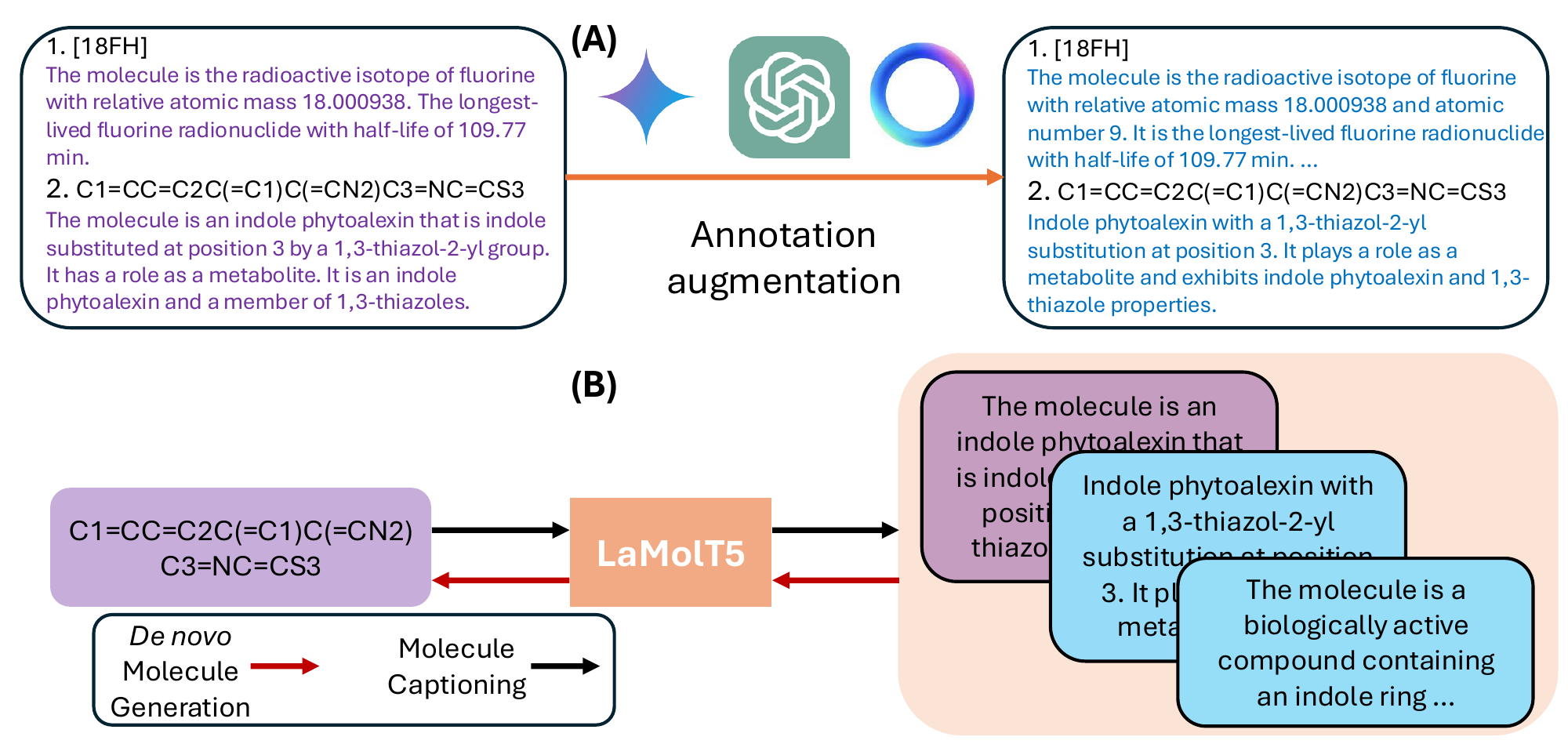}
\caption{
An example implementation of \pipeline for annotation augmentation (A) and training (B).
Given molecules and their \textcolor{violet}{original annotations}, we prompt LLMs to generate \textcolor{oceanboatblue}{augmented annotations} (\newdataset) by rewriting the original annotations. 
Next, we train \newmodel on \newdataset to learn a mapping function between the molecule's SMILES string and corresponding annotations.  
}
\vspace{-3mm}
\label{fig:rewrite_training_pipeline}
\end{figure}

%
To facilitate the practical re-implementation, we showcase details of \pipeline, using \olddataset~\cite{EZJ21}, a widely adopted dataset for molecular generation research (Section~\ref{subsec:preliminary}-\ref{subsec:training}). 
%
In addition, Section~\ref{subsec:extension_broad_application} describes extensive implementations of \pipeline across broad applications in \emph{image}, \emph{text} and \emph{graph} tasks. 

\subsection{Showcasing \olddataset}
\label{subsec:preliminary}


\spara{\olddataset.}
\olddataset contains $33\,010$ molecular entities centred on chemical compounds. 
Each molecule is represented using a SMILES string~\cite{W88} and associated with a high-quality, manually annotated caption supporting various computational and experimental studies.
Given a molecule, we formally represent it as $\mathcal{M} = (S, C)$, where $S$ and $C$ denote its SMILES string and associated caption. 
Examples are illustrated in Figure~\ref{fig:rewrite_training_pipeline}-(A). 
Consequently, \olddataset can be formally represented as $\mathcal{D} = \{ \mathcal{M}_1, \mathcal{M}_2, \dots, \mathcal{M}_n \}$. 
$\mathcal{S} = \{ S_1, S_2, \dots, S_n \}$ represents the SMILES string set and $\mathcal{C} = \{ C_1, C_2, \dots, C_n \}$ the caption (annotation) set.
The dataset is publicly available with a fixed split: $\mathcal{D}_{\textsc{train}}$ ($80\%$), $\mathcal{D}_{\textsc{valid}}$ ($10\%$) and $\mathcal{D}_{\textsc{test}}$ ($10\%$), allowing researchers to consistently train and evaluate their models.

\spara{Tasks.}
\olddataset supports two molecular tasks: \emph{(1)} text-based \emph{de novo} molecule generation (\textsc{gen}) and \emph{(2)} molecule captioning (\textsc{cap}). 
The goal of text-based \emph{de novo} molecule generation is to train a model which can generate a variety of possible new molecules with desired properties as described in the text. 
Specifically, for \olddataset, we aim to learn a model $f_{\textsc{gen}}: \mathcal{C} \to \hat{\mathcal{S}} $ by minimising the loss function value $\min_{\Psi} \mathcal{L}(\hat{\mathcal{S}}_{\textsc{train}}, \mathcal{S}_{\textsc{train}})$, where $\Psi$ represents the set of trainable parameters of $f_{\textsc{gen}}$. 
The target of molecule captioning is to generate descriptions of the components and chemical functionality of a molecule. 
Similarly, we aim to learn a model $f_{\textsc{cap}}: \mathcal{S} \to \hat{\mathcal{C}} $, by minimising loss function value $\min_{\Phi} \mathcal{L}(\hat{\mathcal{C}}_{\textsc{train}}, \mathcal{C}_{\textsc{train}})$, where $\Phi$ represents the set of trainable parameters of $f_{\textsc{cap}}$. 

\spara{\oldmodel.}
The molecule generation tasks can be considered translation tasks, and sequence-to-sequence models serve as solid solutions. 
One fundamental method in this category is \oldmodel~\cite{ELRHCJ22}, an improved version of T5~\cite{RSRLNMZLL20}. 
\oldmodel initialise an encoder-decoder transformer model using public checkpoints of T5. 
The model is then pre-trained on a combined dataset of C4~\cite{RSRLNMZLL20} and ZINC~\cite{SI15} for $1$ million steps.
Finally, it undergoes $50,000$ steps of fine-tuning on \olddataset for two molecular tasks. 
Since these tasks are formulated as sequence-to-sequence tasks, the model is trained using standard maximum likelihood, such as cross-entropy loss  ($\mathcal{L}_{\text{CE}}$). 
Take the fine-tuning phase as an example. 
The parameters are optimised to match the generated text with $\mathcal{D}_{\textsc{train}}$'s text:
\begin{align}
\begin{split}
    \mathcal{L}_{\text{CE}}^{\textsc{gen}} &= - \frac{\sum\limits_{i = 1}^{n} \log p(S_{\textsc{train}, i} \;\vert\; C_{\textsc{train}, i})}{n}, \\
    \mathcal{L}_{\text{CE}}^{\textsc{cap}} &= - \frac{\sum\limits_{i = 1}^{n} \log p(C_{\textsc{train}, i} \;\vert\; S_{\textsc{train}, i})}{n},
\end{split}
\label{eq:molt5}
\end{align}
where $n$ is the number of molecules in $\mathcal{D}_{\textsc{train}}$ and $p(S_{\textsc{train}, i} \;\vert\; C_{\textsc{train}, i})$ is the probability assigned by $f_{\textsc{gen}}$ to the $i$-th true SMILES string $S_{\textsc{train}, i}$ given the $i$-th caption $C_{\textsc{train}, i}$. 
This optimisation increases the probability of generating the correct outputs given the corresponding inputs.
Following this way, \oldmodel provides three trained variants of varying sizes: \oldmodel-Small ($77M$ parameters), \oldmodel-Base ($250M$ parameters) and \oldmodel-Large ($800M$ parameters). 


\subsection{Automatic Annotation Augmentation}
\label{subsec:language_augmentation}

As shown in Equation~\ref{eq:molt5}, the number of training instances directly affects the amount of information injected into the model. 
\citet{ELRHCJ22} discuss the potential limitations caused by the limited data in \olddataset. 
%
A recent breakthrough known as In-Context Learning (ICL) has enhanced the adaptability of LLMs by enabling them to acquire contextual knowledge during inference, eliminating the need for extensive fine-tuning~\cite{CTR20}. 
To harness ICL for \olddataset augmentation, we first formulate a prompt for querying LLMs.
The goal in prompt engineering is to find the correct way to formulate a question $Q$ in such a way that an LLM ($f_{\textsc{LLM}}$) will return the corresponding answer $A$ essentially represented as $A = f_{\textsc{LLM}}(Q)$. 
In this work, we design the prompt as shown in Appendix~\ref{sec:appendix_prompt_details}. 
Precisely, the prompt consists of two components: \emph{Instruction}: Provides general guidance to the LLM, clarifying its role in the conversation; \emph{Message}: Tasks the LLM to rewrite the molecule caption, considering the chemical expertise and given information.

Given an instance $\mathcal{M}_i = (S_i, C_i)$ from $\mathcal{D}_{\textsc{train}}$, we can generate $k$ rewritten captions $\{ C_{i, 1}, C_{i, 2}, \dots, C_{i, k} \}$ with multiple rounds of queries. 
This results in an augmented instance, $\mathcal{M}^{+}_i = (S_i, C_{i, 0}, C_{i, 1}, C_{i, 2} \dots, C_{i, k})$, and an augmented dataset, \newdataset ($\mathcal{D}^{+}$).
Specifically, $\mathcal{D}^{+} = (\mathcal{D}^{+}_{\textsc{train}}, \mathcal{D}_{\textsc{valid}}, \mathcal{D}_{\textsc{test}})$, where $\mathcal{D}^{+}_{\textsc{train}} = \{ \mathcal{M}^+_1, \mathcal{M}^+_2, \dots \}$.
Each SMILES string of $\mathcal{D}^{+}_{\textsc{train}}$ is associated with $k+1$ captions. 
This language augmentation process introduces diversity in sentence structure and vocabulary while preserving the essential knowledge about the molecules.
In our experiments, we adopt two open-source LLMs (Llama 2-70b~\cite{TMSA23} and Llama 3-70b~\cite{TMSA23}) and two closed-source LLMs (GPT 3.5-turbo~\cite{AAAA23} and Gemini Pro~\cite{Gemini24}) to generate four rewritten captions. 
We demonstrate some generated example captions in Table~\ref{table:rewritten_captions} of Appendix~\ref{sec:appendix_caption_details}. 

\subsection{Training on Augmented Dataset}
\label{subsec:training}

After generating $k$ new captions for each molecule of the training dataset $\mathcal{D}_{\textsc{train}}$. 
We proceed to train a model based on \newdataset to perform the molecular tasks, \ie, text-based \emph{de novo} molecule generation (\textsc{gen}) and molecule captioning (\textsc{cap}). 
In this work, we initialise encoder-decoder transformer models using the available \oldmodel variants, as introduced in Section~\ref{subsec:preliminary}. 
We then train a novel model, \newmodel, using a cross-entropy loss:
\begin{align}
\begin{split}
    \mathcal{L}_{\text{CE}}^{\textsc{gen}} &= - \frac{\sum\limits_{i = 1}^{n} \textcolor{orange}{\sum\limits_{j = 1}^{k+1}} \log p(S_{\textsc{train}, i} \vert C^{\textcolor{orange}{+}}_{\textsc{train}, i, \textcolor{orange}{j}})}{\textcolor{orange}{(k+1)}n},\\
    \mathcal{L}_{\text{CE}}^{\textsc{cap}} &= - \frac{\sum\limits_{i = 1}^{n} \textcolor{orange}{\sum\limits_{j = 1}^{k+1}} \log p(C^{\textcolor{orange}{+}}_{\textsc{train}, i, \textcolor{orange}{j}} \vert S_{\textsc{train}, i})}{\textcolor{orange}{(k+1)}n},
\end{split}
\label{eq:lamolt5}
\end{align}
where $p(S_{\textsc{train}, i} \;\vert\; C^+_{\textsc{train}, i, j})$ is the probability assigned by $f_{\textsc{gen}}$ to the $i$-th true SMILES string $S_{\textsc{train}, i}$ given the $i$-th molecule's $j$-th caption $C_{\textsc{train}, i, j}$. 
The critical addition to the \oldmodel is the augmented caption set $C^+_{\textsc{train}, i, \{0, 1, \dots, k \}}$ for each SMILES string $S_{\textsc{train}, i}$. 
For the molecule generation task, \newmodel is trained to generate the correct SMILES string $S_{\textsc{train}, i}$ by giving different caption inputs $C^+_{\textsc{train}, i, \{0, 1, \dots, k \}}$. 
By introducing diversity into the training data, we enhance the effectiveness and robustness of the model in generating SMILES strings (validated in Section~\ref{subsec:molecule_generation}). 
Meanwhile, for the molecule captioning task, \newmodel is trained to generate various captions for each SMILES string. 
Despite these captions having different sentence structures and vocabularies, they preserve the essential knowledge about the molecules.
This training enhances the model’s ability to generate more semantically precise and meaningful captions (validated in Section~\ref{subsec:molecule_captioning}). 
Overall, this novel and effective methodology augments the biomedical dataset without requiring human efforts, significantly contributing to the performance of \newmodel.

\subsection{Extension to Broad Applications}
\label{subsec:extension_broad_application}
To further demonstrate the versatility of \pipeline, we extend its application to several additional datasets, including \bace, \hiv, \esol~\cite{HFZDRLCL20}, and \image~\cite{SDGS18}. 
These datasets support a variety of crucial tasks, \eg, \emph{image captioning}, \emph{text understanding}, and \emph{graph property prediction}.
%
Due to the paper length constraints, the detailed implementations of \newmodel on these additional datasets are presented in Appendix~\ref{sec:appendix_additional_experiments}. 
Experimental results derived from these implementations will be discussed in Section~\ref{subsec:broad_application}. 


\section{Experiments} 
\label{sec:experimental_study}
\subsection{Experimental Settings}
\label{subsec:experimental_settings}

\para{Dataset.}
We use our generated \newdataset dataset for training \newmodel.
One annotation is the original present in \olddataset, and the other two are LLM-generated.
We adopt two conventional LLMs, GPT 3.5-turbo and Gemini Pro, to generate annotations in the main dataset from most experiments. 
We additionally experiment in Section~\ref{subsec:analysis} with annotations generated by two open-source LLMs, \eg, Llama 2-70b and Llama 3-70b. 

\spara{Baselines.}
We mainly consider three families of methods: 
\textbf{(1)} Methods included in the benchmark paper~\cite{ELRHCJ22}, including RNN~\cite{CMGBBSB14}, Transformer~\cite{VSPUJGKP17}, T5~\cite{RSRLNMZLL20} and \oldmodel~\cite{ELRHCJ22}. 
\textbf{(2)} The state-of-the-art methods (reported on the leaderboards~\cite{denovo,captioning}) without additional datasets, including Text+Chem T5~\cite{CGBWLM23}, TGM-DLM~\cite{GLWW24} and MolReGPT~\cite{LLFWLTL24}. 
They rely on the same datasets as \newmodel, we report their results in Table~\ref{table:results_mol_generation}, \ref{table:results_mol_caption}
\textbf{(3)} The state-of-the-art methods incorporating extra datasets. 
For the text-based \emph{de novo} molecule generation task, we include BioT5~\cite{PZZWGWXY23} and BioT5+~\cite{PWGLFZXQY24}; for the molecule captioning task, we include BioT5, MolXPT~\cite{LZXWXQZL23}. 
For a fair comparison, we report their results in Figure~\ref{fig:compare_sota}. 
A detailed description of each baseline method can be found in Appendix~\ref{sec:appendix_baseline}.

\spara{Training Setup.}
We train \newmodel-Small and -Base for as little as $1500$ epochs and \newmodel-Large for $200$ epochs. 
This project used $\sim11500$ H100 GPU hours. 
Detailed hyper-parameter settings and checkpoints are available online\footnote{The augmented dataset and trained models are available at \url{https://anonymous.4open.science/r/LaMolT5-D3C3}.}. 

\spara{Evaluation Setup.}
We adopt the benchmark evaluation setup following~\cite{ELRHCJ22}. 
For the text-based \emph{de novo} molecule generation task, we employ: SMILES \emph{BLEU} score~\cite{PRWZ02}, \emph{Levenshtein} distance~\cite{MVM09}, Fréchet ChemNet Distance (\emph{FCD})~\cite{PRUHK18}, \emph{MACCS FTS}~\cite{DLHN02}, \emph{RDK FTS}~\cite{SSL15} \emph{Morgan FTS}~\cite{RH10}, \emph{Exact} score~\cite{ELRHCJ22}, \emph{Validity}~\cite{ELRHCJ22}, and \emph{Text2Mol}~\cite{EZJ21}. 
For the molecule captioning task, we measure the \emph{BLEU} score, \emph{ROUGE}~\cite{L04}, \emph{METEOR}~\cite{BL05}, and \emph{Text2Mol} of the generated annotation compared to the ground-truth. 
Detailed descriptions can be found in Appendix~\ref{sec:appendix_evaluation}.

\subsection{Molecule Generation}
\label{subsec:molecule_generation}

\begin{table*}[!ht]
\centering
\small
\resizebox{1.\linewidth}{!}{
\begin{tabular}{@{}llllllllll@{}}
    \toprule
    Model & BLEU$\uparrow$ & Exact$\uparrow$ & Levenshtein$\downarrow$ & MACCS FTS$\uparrow$ & RDK FTS$\uparrow$ & Morgan FTS$\uparrow$ & FCD$\downarrow$ & Text2Mol$\uparrow$ & Validity$\uparrow$ \\ 
    \midrule
    Ground Truth & 1.000 & 1.000 & 0.0 & 1.000 & 1.000 & 1.000 & 0.0 & 0.609 & 1.0 \\ 
    \midrule
    RNN & 0.652 & 0.005 & 38.09 & 0.591 & 0.400 & 0.362 & 4.34 & 0.409 & 0.542 \\
    Transformer & 0.499 & 0.000 & 57.66 & 0.480 & 0.320 & 0.217 & 16.03 & 0.277 & 0.906 \\ 
    \midrule
    T5-Small & 0.741 & 0.064 & 27.703 & 0.704 & 0.578 & 0.525 & 1.77 & 0.479 & 0.608 \\
    MolT5-Small & 0.755 & 0.079 & 25.988 & 0.703 & 0.568 & 0.517 & 1.35 & 0.482 & 0.721 \\ 
    \textbf{\newmodel-Small} & \textbf{0.852} & \textbf{0.287} & \textbf{16.009} & \textbf{0.891} & \textbf{0.805} & \textbf{0.741} & \textbf{0.384} & \textbf{0.594} & \textbf{0.950} \\
    (Improvement) & +12.85\% & +263.29\% & +38.42\% & +26.72\% & +41.55\% & +43.33\% & +71.56\% & +23.24\% & +31.74\% \\
    \midrule
    T5-Base & 0.762 & 0.069 & 24.950 & 0.731 & 0.605 & 0.545 & 1.43 & 0.499 & 0.660 \\
    MolT5-Base & 0.769 & 0.081 & 24.458 & 0.721 & 0.588 & 0.529 & 1.16 & 0.496 & 0.772 \\ 
    \textbf{\newmodel-Base} & \underline{\textbf{0.861}} & \textbf{0.325} & \underline{\textbf{14.685}} & \textbf{0.899} & \underline{\textbf{0.819}} & \underline{\textbf{0.760}} & \underline{\textbf{0.352}} & \underline{\textbf{0.596}} & \textbf{0.961} \\
    (Improvement) & +11.96\% & +301.23\% & +39.95\% & +24.69\% & +39.29\% & +43.67\% & +69.66\% & +20.16\% & +24.48\% \\
    \midrule
    T5-Large & 0.854 & 0.279 & 16.721 & 0.823 & 0.731 & 0.670 & 0.401 & 0.552 & 0.902 \\
    MolT5-Large & 0.854 & 0.311 & 16.071 & 0.834 & 0.746 & 0.684 & 0.385 & 0.554 & 0.905 \\
    \textbf{\newmodel-Large} & \textbf{0.856} & \underline{\textbf{0.328}} & \textbf{15.666} & \textbf{0.892} & \textbf{0.816} & \textbf{0.754} & \textbf{0.371} & \textbf{0.593} & \underline{\textbf{0.962}} \\
    \text{(Improvement)} & +0.23\% & +5.46\% & +2.52\% & +6.9\% & +9.38\% & +10.23\% & +3.64\% & +7.04\% & +6.30\% \\
    \midrule
    Text+Chem T5 & 0.750 & 0.212 & 27.39 & 0.874 & 0.767 & 0.697 & 0.499 & 0.574 & 0.792 \\
    TGM-DLM & 0.826 & 0.242 & 17.00 & 0.854 & 0.739 & 0.688 & 0.770 & 0.581 & 0.871 \\
    MolReGPT & 0.857 & 0.280 & 17.14 & \underline{0.903} & 0.805 & 0.739 & 0.410 & 0.593 & 0.899 \\
    \bottomrule
\end{tabular}
}
\caption{
Text-based \emph{de novo} molecule generation results for models without additional datasets. 
Models incorporating extra datasets are presented in Figure~\ref{fig:compare_sota}. 
Best performances are highlighted with an \underline{underline}.
}
\vspace{-3mm}
\label{table:results_mol_generation}
\end{table*}

\begin{table*}[!ht]
\centering
\resizebox{.8\linewidth}{!}{
\begin{tabular}{>{\arraybackslash}p{10cm} p{3cm} p{3cm} p{3cm}} 
    \toprule
    \textbf{Caption} & \textbf{Ground Truth} & \textbf{\oldmodel-Small} & \textbf{\newmodel-Small} \\ 
    \midrule
    \textbf{(1): }
    The molecule is an \textcolor{violet}{indolylmethylglucosinolate} that is the conjugate base of 4-methoxyglucobrassicin, obtained by deprotonation of the sulfo group. It is a conjugate base of a 4-methoxyglucobrassicin. & \raisebox{-\totalheight}{\includegraphics[width=3cm,height=3cm]{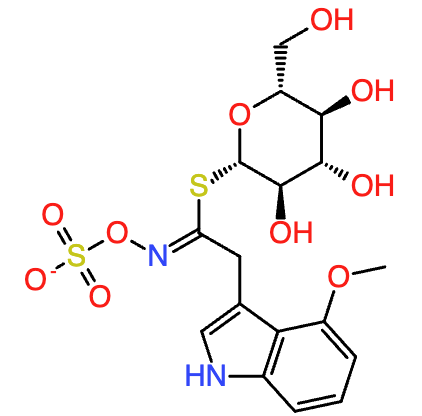}} & \raisebox{-\totalheight}{\includegraphics[width=3cm,height=2.5cm]{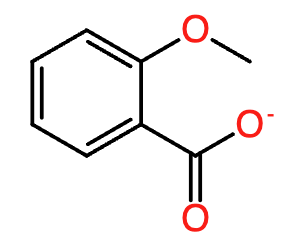}} & \raisebox{-\totalheight}{\includegraphics[width=3cm,height=3cm]{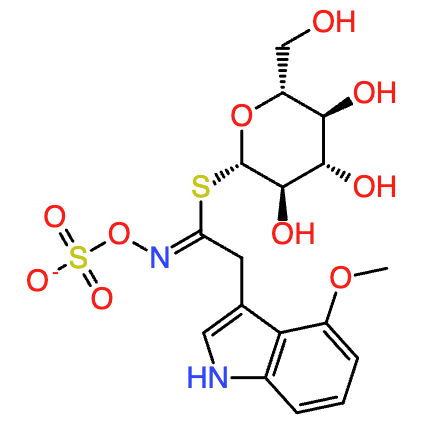}} \\ 
    \midrule
    \textbf{(2): }
    The molecule is an \textcolor{violet}{N-carbamoylamino acid} that is aspartic acid with one of its \textcolor{violet}{amino hydrogens} \textcolor{oceanboatblue}{replaced} by a \textcolor{violet}{carbamoyl} group. It has a role as a Saccharomyces cerevisiae metabolite, an Escherichia coli metabolite and a human metabolite. It is a N-carbamoyl-amino acid, an aspartic acid derivative and a C4-dicarboxylic acid. It is a conjugate acid of a N-carbamoylaspartate(2-). & \raisebox{-\totalheight}{\includegraphics[width=3cm,height=2.cm]{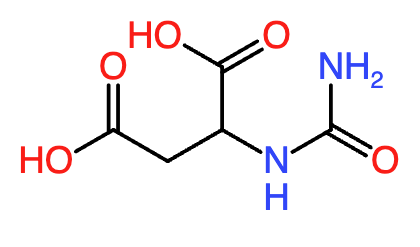}} & \raisebox{-\totalheight}{\includegraphics[width=3cm,height=1.5cm]{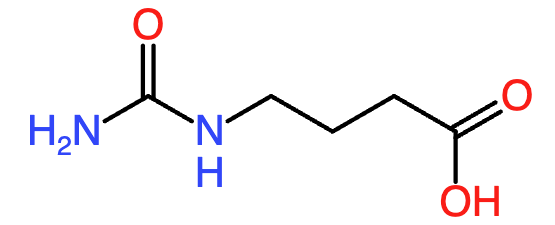}} & \raisebox{-\totalheight}{\includegraphics[width=3cm,height=2.cm]{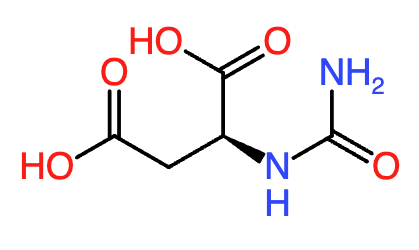}} \\ 
    \bottomrule 
\end{tabular}
}
\caption{
Example of molecules generated with the same input descriptions. 
Generated SMILES strings are converted to molecule graphs for better visualisation.
}
\vspace{-3mm}
\label{table:example_generated_molecule}
\end{table*}

\para{Observation 1: \newmodel significantly elevates the performance of \oldmodel.} 
The \emph{de novo} molecule generation results across nine evaluation metrics in Table~\ref{table:results_mol_generation} reveal that \newmodel achieves the highest performance on all measures.
In addition, \newmodel consistently delivers substantial enhancements over \oldmodel, with improvements up to $301\%$ in terms of Exact score, which measures the number of times the output corresponds to the ground truth. 
This consistent and notable performance underscores the effectiveness of \newdataset and \newmodel. 
Moreover, it illustrates the efficacy of our automatic annotation-augmentation pipeline, \pipeline, on biomedical datasets.

\spara{Observation 2: The small $77M$ parameters \newmodel  outperforms the $800M$ \oldmodel.}
The results of Table~\ref{table:results_mol_generation} indicate that the small-size variant of \newmodel ($77M$) outperforms the large-size variant of \oldmodel ($800M$) in seven different evaluation metrics (Levenshtein, MACCS FTS, RDK FTS, Morgan FTS, FCD, Text2Mo, Validity) on the molecule generation task.
On the other two evaluation metrics (BLEU and Exact), \newmodel-Small achieves competitive performance compared to \oldmodel-Large. 
\newmodel achieves impressive results by leveraging the annotation-augmented dataset, \newdataset, which introduces diversity in sentence structure and vocabulary while maintaining the core molecular knowledge. 

\subsection{Molecule Captioning}
\label{subsec:molecule_captioning}

\begin{table*}[!ht]
\centering
\resizebox{.9\linewidth}{!}{
\begin{tabular}{@{}llllllll@{}}
    \toprule
    Model & BLEU-2$\uparrow$ & BLEU-4$\uparrow$ & ROUGE-1$\uparrow$ & ROUGE-2$\uparrow$ & ROUGE-L$\uparrow$ & METEOR$\uparrow$ & Text2Mol$\uparrow$ \\ 
    \midrule
    Ground Truth & 1.000 & 1.000  & 1.000 & 1.000 & 1.000 & 1.000 & 0.609 \\ 
    \midrule
    RNN & 0.251 & 0.176 & 0.450 & 0.278 & 0.394 & 0.363 & 0.426 \\
    Transformer & 0.061 & 0.027 & 0.204 & 0.087 & 0.186 & 0.114 & 0.057 \\ 
    \midrule
    T5-Small & 0.501 & 0.415 & 0.602 & 0.446 & 0.545 & 0.532 & 0.526 \\
    MolT5-Small & 0.519 & 0.436 & \textbf{0.620} & \textbf{0.469} & \textbf{0.563} & 0.551 & 0.540 \\ 
    \textbf{\newmodel-Small} & \textbf{0.539} & \textbf{0.446} & 0.610 & 0.446 & 0.538 & \textbf{0.566} & \textbf{0.588} \\ 
    (Improvement) & +3.85\% & +2.29\% & -1.61\% & -4.90\% & -4.44\% & +2.72\% & +8.89\% \\
    \midrule
    T5-Base & 0.511 & 0.423 & 0.607 & 0.451 & 0.550 & 0.539 & 0.523 \\
    MolT5-Base & 0.540 & 0.457 & \textbf{0.634} & \textbf{0.485} & \textbf{0.578} & 0.569 & 0.547 \\ 
    \textbf{\newmodel-Base} & \textbf{0.574} & \textbf{0.485} & \textbf{0.634} & 0.478 & 0.564 & \textbf{0.596} & \underline{\textbf{0.599}} \\
    (\text{Improvement}) & +6.30\% & +6.13\% & 0.00\% & -1.44\% & -2.42\% & +4.75\% & +9.51\% \\ 
    \midrule
    T5-Large & 0.558 & 0.467 & 0.630 & 0.478 & 0.569 & 0.586 & 0.563 \\
    MolT5-Large & 0.594 & 0.508 & 0.654 & 0.510 & 0.594 & 0.614 & 0.582 \\
    \textbf{\newmodel-Large} & \textbf{0.602} & \textbf{0.521} & \underline{\textbf{0.655}} & \underline{\textbf{0.512}} & \underline{\textbf{0.598}} & \underline{\textbf{0.634}} & \textbf{0.597} \\
    (\text{Improvement}) & +1.35\% & +2.56\% & +0.15\% & +0.39\% & +0.67\% & +3.26\% & +2.58\% \\
    \midrule
    Text+Chem T5 & 0.580 & 0.490 & 0.647 & 0.498 & 0.586 & 0.604 & 0.567 \\
    MolReGPT & \underline{0.607} & \underline{0.525} & 0.634 & 0.476 & 0.562 & 0.610 & 0.585 \\
    \bottomrule
\end{tabular}
}
\caption{ 
Molecule captioning results for models without additional datasets. 
Models incorporating extra datasets are presented in Figure~\ref{fig:compare_sota}. 
Best performances are highlighted with an \underline{underline}.
}
\label{table:results_mol_caption}
\end{table*}

\begin{table*}[!ht]
\centering
\resizebox{.8\linewidth}{!}{
\begin{tabular}{>{\arraybackslash}p{.5cm} p{3.2cm} p{6.cm} p{6.cm} p{6.cm}} 
    \toprule
    \textbf{ID} & \textbf{Molecule} & \textbf{Ground Truth} & \textbf{\oldmodel-Small} & \textbf{\newmodel-Small} \\ 
    \midrule
    \textbf{1} & \raisebox{-\totalheight}{\includegraphics[width=3.2cm,height=3.2cm]{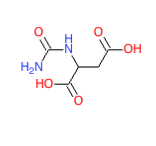}} & The molecule is an N-carbamoylamino acid that is aspartic acid with one of its amino hydrogens replaced by a carbamoyl group. It has a role as a Saccharomyces cerevisiae metabolite, \textcolor{violet}{an Escherichia coli metabolite and a human metabolite}. It is a N-carbamoyl-amino acid, an aspartic acid derivative and a C4-dicarboxylic acid. It is a conjugate acid of a N-carbamoylaspartate(2-). & The molecule is a member of the class of ureas that is urea in which one of the amino hydrogens is replaced by a carbamoyl group. It has a role as a metabolite. It is a N-acyl-amino acid and a member of ureas. & The molecule is a member of the class of ureas that is malonic acid in which one of the hydrogens attached to the nitrogen is substituted by a carbamoyl group. It has a role as \textcolor{violet}{an Escherichia coli metabolite and a mouse metabolite}. It is a member of ureas and a member of ureas. It derives from a malonic acid. It is a conjugate acid of a N-carbamoylglycinate. \\ 
    \midrule
    \textbf{2} & \raisebox{-\totalheight}{\includegraphics[width=3.2cm,height=3.2cm]{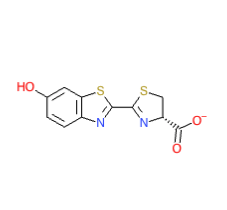}} & The molecule is the conjugate base of Photinus luciferin; major species at pH 7.3. It is a conjugate base of a Photinus luciferin. It is an enantiomer of an ent-Photinus luciferin(1-). & The molecule is a monocarboxylic acid anion that is the conjugate base of ent-Photinus luciferin, obtained by deprotonation of the carboxy group; major species at pH 7.3. It is a conjugate base of an ent-Photinus luciferin. It is an enantiomer of a Photinus luciferin(1-). & The molecule is a monocarboxylic acid anion that is the conjugate base of Photinus luciferin, obtained by deprotonation of the carboxy group; major species at pH 7.3. It is a conjugate base of a Photinus luciferin. It is an enantiomer of a Photinus luciferin(1-). \\ 
    \bottomrule    
\end{tabular}
}
\caption{
Example of captions generated with the same input SMILES strings. 
Input SMILES strings are converted to molecule graphs for better visualisation.
}
\label{table:example_generated_caption}
\vspace{-3mm}
\end{table*}

\spara{Observation 3: \newmodel generates coherent descriptions.}
The results in Table~\ref{table:results_mol_caption} highlight the superior performance of the \newmodel in the molecule captioning task. 
\newmodel excels in the Text2Mol metric, which provides a comprehensive assessment of the \textit{semantic alignment} ---throuch cosine similarity-- between generated descriptions and their corresponding molecules.
\newmodel variants achieve improvements up to $23\%$ in \newmodel-Small over the corresponding \oldmodel variants. 
These results underscore \newmodel’s enhanced ability to capture the intricate semantics of molecule descriptions, making it a highly effective model for this task.

\spara{Observation 4: \newmodel exhibits lower ROUGE score.}
\newmodel exhibits lower ROUGE scores than \oldmodel, as ROUGE emphasises exact n-gram overlaps, which do not fully capture the semantic accuracy of the generated text. 
The \newdataset, by introducing variability in sentence structure and vocabulary, contributes to this discrepancy. 
During training, \newmodel prioritises achieving higher semantic coherence, potentially at the expense of exact word or phrase matches. 
As we argue above, although \newmodel excels in capturing the overall semantics of molecule descriptions, this focus leads to reduced performance on ROUGE scores.

\subsection{Analysis}
\label{subsec:analysis}

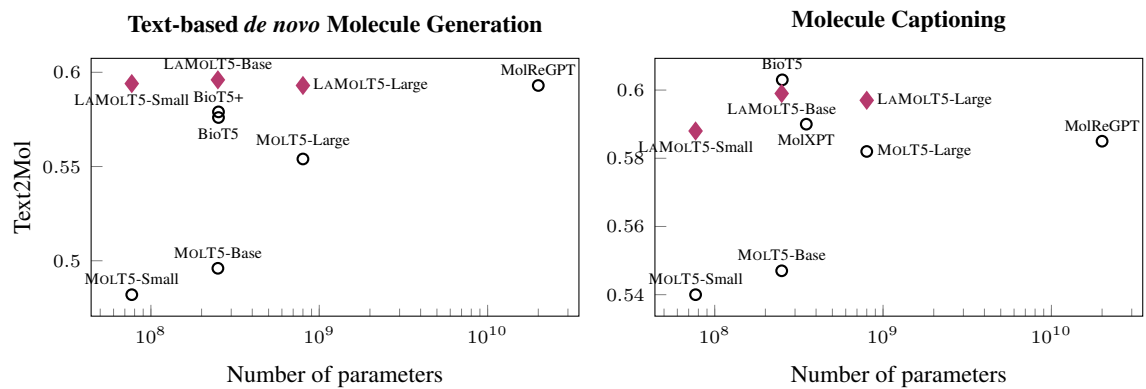
\begin{figure*}[!ht]
\begin{tikzpicture}
\begin{groupplot}[
    group style={
    group size= 2 by 1},
    enlargelimits=0.1, 
    xmode=log, 
    xlabel={Number of parameters},
    xlabel style={font=\small}, 
    width=.5\textwidth, 
    height=5cm, 
    xtick pos=left,
    ytick pos=left,
    xticklabel style={font=\scriptsize},
    yticklabel style={font=\scriptsize},
]
    \nextgroupplot[
        title={\textbf{Text-based \textit{de novo} Molecule Generation}},
        title style={font=\small},
        ylabel={Text2Mol}, 
        ylabel style={font=\small, at={(0.05,0.5)}}, 
        scatter/classes={c={mark=diamond*, mark size=3pt, thick, cycle4!60!cycle1}, n={mark=o, mark size=2pt, black, thick}},
        scatter, 
        mark=*, 
        only marks, 
        scatter src=explicit symbolic,
    ]
    \addplot[
        nodes near coords style = {anchor=south},
        nodes near coords*={\Label},
        visualization depends on={value \thisrow{model} \as \Label},
        every node/.append style ={font=\tiny}
    ] table [x=x, y=y, meta = class, col sep=comma] {
        x,y,model,class
        250000000.0,0.596,\newmodel-Base,c
        77000000.0,0.482,\oldmodel-Small,n
        250000000.0,0.496,\oldmodel-Base,n
        800000000.0,0.554,\oldmodel-Large,n
        20000000000.0,0.593,MolReGPT,n
        252000000.0,0.579,BioT5+,n
    };
    \addplot[
        nodes near coords style = {anchor=north },
        nodes near coords*={\Label},
        visualization depends on={value \thisrow{model} \as \Label},
        every node/.append style ={font=\tiny}
    ] table [x=x, y=y, meta = class, col sep=comma] {
        x,y,model,class
    77000000.0,0.594,\newmodel-Small,c
    };
    \addplot[
        nodes near coords style = {anchor=west},
        nodes near coords*={\Label},
        visualization depends on={value \thisrow{model} \as \Label},
        every node/.append style ={font=\tiny}
    ] table [x=x, y=y, meta = class, col sep=comma] {
        x,y,model,class
        800000000.0,0.593,\newmodel-Large,c
    };
    \addplot[
        nodes near coords style = {anchor=north },
        nodes near coords*={\Label},
        visualization depends on={value \thisrow{model} \as \Label},
        every node/.append style ={font=\tiny}
    ] table [x=x, y=y, meta = class, col sep=comma] {
        x,y,model,class
    252000000.0,0.576,BioT5,n
    };
    
    \nextgroupplot[
        title={\bf{Molecule Captioning}}, 
        title style={font=\small},
        scatter/classes={c={mark=diamond*, mark size=3pt, thick, cycle4!60!cycle1}, n={mark=o, mark size=2pt, black, thick}},
        scatter, 
        mark=*, 
        only marks, 
        scatter src=explicit symbolic,
    ]
    \addplot[
        nodes near coords style = {anchor=south},
        nodes near coords*={\Label},
        visualization depends on={value \thisrow{model} \as \Label},
        every node/.append style ={font=\tiny}, 
    ] table [x=x, y=y, meta = class, col sep=comma] {
        x,y,model,class
        77000000.0,0.54,\oldmodel-Small,n
        250000000.0,0.547,\oldmodel-Base,n
        20000000000.0,0.585,MolReGPT,n
        252000000.0,0.603,BioT5,n
    };
    \addplot[
        nodes near coords style = {anchor=west},
        nodes near coords*={\Label},
        visualization depends on={value \thisrow{model} \as \Label},
        every node/.append style ={font=\tiny}, 
    ] table [x=x, y=y, meta = class, col sep=comma] {
        x,y,model,class
        800000000.0,0.597,\newmodel-Large,c
    };
    \addplot[
        nodes near coords style = {anchor=west},
        nodes near coords*={\Label},
        visualization depends on={value \thisrow{model} \as \Label},
        every node/.append style ={font=\tiny}, 
    ] table [x=x, y=y, meta = class, col sep=comma] {
        x,y,model,class
        800000000.0,0.582,\oldmodel-Large,n
    };
    \addplot[
        nodes near coords style = {anchor=north },
        nodes near coords*={\Label},
        visualization depends on={value \thisrow{model} \as \Label},
        every node/.append style ={font=\tiny}
    ] table [x=x, y=y, meta = class, col sep=comma] {
        x,y,model,class
    350000000.0,0.590,MolXPT,n
    };
    \addplot[
        nodes near coords style = {anchor=north },
        nodes near coords*={\Label},
        visualization depends on={value \thisrow{model} \as \Label},
        every node/.append style ={font=\tiny}
    ] table [x=x, y=y, meta = class, col sep=comma] {
        x,y,model,class
    250000000.0,0.599,\newmodel-Base,c
    };
    \addplot[
        nodes near coords style = {anchor=north },
        nodes near coords*={\Label},
        visualization depends on={value \thisrow{model} \as \Label},
        every node/.append style ={font=\tiny}
    ] table [x=x, y=y, meta = class, col sep=comma] {
        x,y,model,class
    77000000.0,0.588,\newmodel-Small,c
    };
    
\end{groupplot}
\end{tikzpicture}
\caption{
Performance vs. Number of parameters of \newmodel and top-$3$ leaderboard state-of-the-art methods. 
Overall rank: \newmodel-{Base} (\#1), \newmodel-{Large} (\#2) and BioT5 (\#3). 
}
\vspace{-3mm}
\label{fig:compare_sota}
\end{figure*}

\para{Comparison with state-of-the-art methods.}
Figure~\ref{fig:compare_sota} demonstrates the performance of top-$3$ leaderboard SOTA models.
\newmodel variants take the first two overall ranks. 
Notably, \newmodel establishes new SOTA results on molecule generation task. 
Although BioT5 outperforms \newmodel on the molecule captioning task, \emph{BioT5 leverages external data}, offering an additional advantage. 
When comparing models that do not incorporate external knowledge, \newmodel achieves the best performance, solidifying its position as the top model in this domain. 
Moreover, the small-size variant delivers highly competitive results with significantly fewer parameters than other leading models: \newmodel-Small has $99\%$ fewer parameters than MolRecGPT but delivers superior performance. 
This efficiency makes \newmodel-Small an attractive option for applications requiring high performance with reduced computational resources.

\spara{Performance with different augmentations.} 
Figure~\ref{fig:ablation} shows the performance of \newmodel with different augmentation strategies during training. 
\emph{(i)} The combination of two augmented annotations generated by \pipeline demonstrates a consistent improvement to using only one augmentation in performance, leading to more robust learning. 
\emph{(ii)} Conventional data augmentation strategies, \eg, EDA~\cite{WZ19}, Mixup~\cite{ZCDL18} are not feasible solutions in the context of biological data. 
We argue that the pre-training stage involved general knowledge about molecules, and easy text wrapping does not provide enough diversity to enhance the information. 
\emph{(iii)} Relying on captions generated directly by LLMs not only fails to improve performance but actually degrades it.
Specifically, including these directly generated annotations results in lower BLEU scores than the original \oldmodel, underscoring that such annotations might introduce noise or lack the molecular knowledge needed for effective training.
Such results echo the empirical studies of \cite{ZZM24} and affirm the necessity of \pipeline design. 

\pgfplotstableread[col sep = comma]{./figures/mol-gen-data.csv}\data

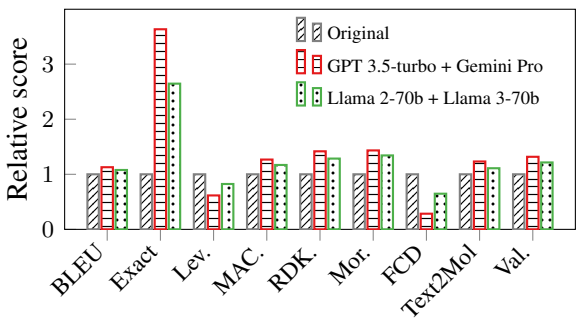
\begin{figure}[!ht]
\centering
\begin{tikzpicture}
\begin{axis}[
    height=4.5cm,
    ymin=0,
    xticklabels={BLEU, Exact, Lev., MAC., RDK., Mor., FCD, Text2Mol, Val.},
    ybar=1pt, 
    x=0.7cm, 
    bar width=0.15cm, 
    every axis plot/.append style={thick},
    scaled ticks=false,
    xtick=data,
    ylabel={Relative score},
    ylabel style={at={(0.09, 0.5)}},
    legend cell align={left},
    legend pos=north west,
    legend columns=1,
    legend style={font=\small,at={(0.7,0.5)},anchor=south, font=\scriptsize, draw=none},
    legend entries={Original, GPT 3.5-turbo + Gemini Pro, Llama 2-70b + Llama 3-70b},
    xlabel style={align=center}, 
    ytick pos=left,
    xtick pos=left,
    x tick label style={font=\small,rotate=45,anchor=east},
    y tick label style={font=\small}, 
]

\addplot[gray, fill=gray!50, pattern=north east lines] table [x expr=\coordindex, y={Original}]{\data};
\addplot[cycle1, fill=cycle1!50, pattern=horizontal lines] table [x expr=\coordindex, y={GPT 3.5-turbo + Gemini Pro}]{\data};
\addplot[cycle3, fill=cycle3!50, pattern=dots] table [x expr=\coordindex, y={Llama 2-70b + Llama 3-70b}]{\data};

\end{axis}
\end{tikzpicture}
\caption{Molecule generation performance of \oldmodel-Small and \newmodel-Small with captions generated by open-sourced and closed-sources LLMs.}
\vspace{-3mm}
\label{fig:performance_diff_llms}
\end{figure}

\spara{Performance of \pipeline using different LLMs.}
Figure~\ref{fig:performance_diff_llms} show the molecule generation performance based on annotations augmented by different LLMs. 
\newmodel trained on \newdataset annotations generated by both open- and closed-source LLMs consistently outperforms \oldmodel. 
This highlights the versatility of our proposed annotation augmentation pipeline, \pipeline, in practical applications. 
Expectedly, \newmodel trained on closed-source LLMs (GPT 3.5-turbo + Gemini Pro) outperforms the one trained on open-source LLMs (Llama 2-70b + Llama 3-70b). 
As LLMs continue to improve in performance and in-context learning capabilities, \newmodel can benefit directly from these advancements. 

\subsection{Broad Applications}
\label{subsec:broad_application}

\begin{table}[!ht]
\centering
\small
\begin{tabular}{@{}lllll@{}}
    \toprule
    \multirow{2}{*}{\specialcell{\textbf{Data}\\ \textbf{Eval. Metric}}} & \multirow{2}{*}{\textbf{Task}} & \multirow{2}{*}{\textbf{Model}} & \multicolumn{2}{c}{\textbf{Performance}} \\ 
     & & & Orig. & \pipeline \\ 
    \midrule
    \specialcell{\bace\\ROC-AUC $\uparrow$} & Class. & \specialcell{LM\\GNN} & \specialcell{0.6163\\0.7147 } & \specialcell{\textbf{0.6589} \\ \textbf{0.7760}} \\ 
    \midrule
    \specialcell{\hiv\\ROC-AUC $\uparrow$} & Class. & \specialcell{LM\\GNN} & \specialcell{0.5037 \\ 0.7376} & \specialcell{\textbf{0.5562} \\ \textbf{0.7641}} \\ 
    \midrule
    \specialcell{\esol\\RMSE $\downarrow$} & Reg. & \specialcell{LM\\GNN} & \specialcell{2.2549 \\ 1.2561} & \specialcell{\textbf{2.1811} \\ \textbf{0.9301}} \\ 
    \midrule
    \specialcell{\image\\Accuracy $\uparrow$} & Class. & CNN & 15.8 & \textbf{17.7} \\
    \bottomrule
\end{tabular}
\caption{
Results on \emph{image}, \emph{text} and \emph{graph} tasks. 
LM: DeBERTa~\cite{HGC23};
GNN: GCN~\cite{KW17};
CNN: ViT-B/16~\cite{DBKWZUDMHGUH21}. 
}
\label{table:results_more_dataset}
\vspace{-3mm}
\end{table}

\spara{Results of \emph{image}, \emph{text} and \emph{graph} tasks.}
To further demonstrate the versatility of \pipeline, we perform extended experiments on several additional datasets, including \bace, \hiv, \esol, and \image, which support a variety of crucial tasks, \eg, \emph{image captioning}, \emph{text understanding}, and \emph{graph property prediction}.
Due to the page limit, we describe the detailed implementations in Appendix~\ref{sec:appendix_additional_experiments}. 
Results in Table~\ref{table:results_more_dataset} show that \pipeline significantly enhances performance across these diverse applications. 
This improvement highlights \pipeline’s potential to be a valuable tool in a wide range of AI tasks, offering substantial gains in accuracy and efficiency.


\section{Concluding Discussion} 
\label{sec:conclusion}
This work proposes an automatic annotation augmentation pipeline, \pipeline, designed to enhance annotated datasets and thereby boost the performance of AI approaches with minimal cost. 
We generate \newdataset, an enriched biomedical dataset featuring diverse sentences and vocabulary while preserving essential molecular knowledge.
This increased diversity is crucial for training \newmodel models, leading to remarkable improvements in challenging molecular tasks. 
A set of ablation studies investigate the impact of \pipeline design and affirm its effectiveness. 
Furthermore, we expand the application of \pipeline to a wide range of datasets across different domains, including \emph{image captioning}, \emph{text understanding}, and \emph{graph property prediction}.
The observed significant improvements vindicate the remarkable versatility and utility of \pipeline.


\section{Limitations and Ethic Statement} 
\label{sec:limitations}
\spara{Limitations.}
The language augmentation process relies on external LLMs, which introduce uncertainties because their robustness in other applications cannot be guaranteed. 
Additionally, results shown in Figure~\ref{fig:performance_diff_llms} demonstrate the impact of caption quality on molecular tasks, suggesting that developing techniques for filtering captions could be a valuable direction for future work.
Moreover, while LLMs continue to improve in performance and ICL capabilities, \newmodel can benefit from these advancements. 
However, the domain-specific knowledge embedded in LLMs remains relatively limited. 
Thus, exploring practical solutions to incorporate more comprehensive domain knowledge into LLMs for language augmentation is a promising future direction for enhancing \pipeline.

\spara{Ethic Statement.} 
Throughout our work, we did not utilise any private or sensitive information.
The involved datasets are open-source, and outputs are available online to the community. 
However, it's essential to note that if any private information were to be inadvertently exposed to an LLM during internal pre-training and fine-tuning stages, \pipeline does not offer any privacy filtration mechanism.
Therefore, there exists the potential for privacy concerns associated with the underlying model to manifest through the output provided by \pipeline. 


\section*{Acknowledgments}
Zhiqiang Zhong is partially funded by the Aarhus University Research Foundation (AUFF) under the grant GAI (2022-0142).

\bibliography{full_format_references}

\newpage
\appendix
\section{Proposed Prompt Details}
\label{sec:appendix_prompt_details}

In this work, we design the prompt as follows:
\begin{mybox}
    \textbf{Instruction: }
    You are now a chemical specialist in rewriting captions for a molecule in SMILES format. Make sure those captions describe the given molecule correctly and precisely based on your two inputs (SMILES and Caption of it). 
    Also, make sure your rewriting captions do not include the input SMILES.
    Write the response without using linebreaks, newlines, or special characters such as ``\textbackslash t'' or ``\textbackslash n''.
    \\
    \textbf{Message: }
    \emph{SMILES string of target molecule:} [SMILES]. \\
    \emph{Description of the molecule:} [caption]. \\
    \emph{Task:} Rewrite the following molecule with its SMILES and caption. 
    The newly rewritten caption should be elaborate, descriptive, and concise, highlighting the key structural features and biological activities of the molecule. Only output rewritten caption without any header and linebreak. \\
    \emph{Answer:}
\end{mybox}

\section{Caption Augmentation Details} 
\label{sec:appendix_caption_details}

\begin{table*}[!ht]
\centering
\resizebox{1.\linewidth}{!}{
\begin{tabular}{>{\arraybackslash}p{3cm} p{5cm} p{5cm} p{5cm}} 
    \toprule
    \textbf{Molecule} & \textbf{Caption - Original} & \textbf{Caption - GPT 3.5-turbo} & \textbf{Caption - Gemini Pro} \\ 
    \midrule 
    \raisebox{-\totalheight}{\includegraphics[width=2.5cm,height=2.5cm]{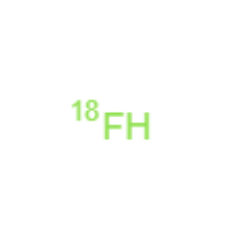}} & \small The molecule is the radioactive isotope of fluorine with relative atomic mass 18.000938. The longest-lived fluorine radionuclide with half-life of 109.77 min. & \small The molecule is the radioactive isotope of fluorine with relative atomic mass 18.000938 and atomic number 9. It is the longest-lived fluorine radionuclide with half-life of 109.77 min. It is used in positron emission tomography (PET) for imaging the brain, heart, and other organs. & \small This molecule is a radioactive isotope of fluorine with a relative atomic mass of 18.000938. It is the longest-lived fluorine radionuclide with a half-life of 109.77 minutes. \\ 
    \midrule
    \raisebox{-\totalheight}{\includegraphics[width=2.5cm,height=3cm]{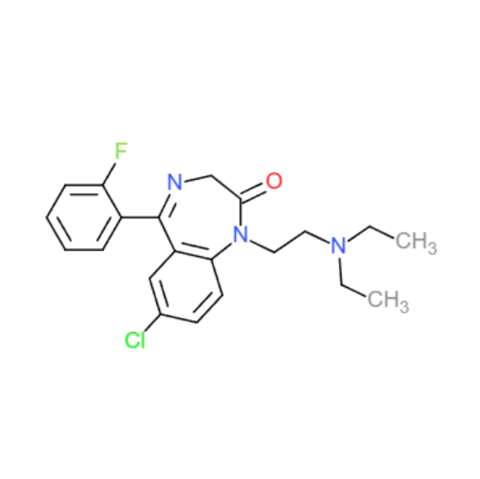}} & \small The molecule is a 1,4-benzodiazepinone that is 1,3-dihydro-2H-1,4-benzodiazepin-2-one substituted by a 2-(diethylamino)ethyl group, 2-fluorophenyl group and chloro group at positions 1, 5 and 7, respectively. It is a partial agonist of GABAA receptors and used for the treatment of insomnia. It has a role as a sedati gve, an anticonvulsant, a GABAA receptor agonist and an anxiolytic drug. It is a 1,4-benzodiazepinone, an organochlorine compound, a member of monofluorobenzenes and a tertiary amino compound. & \small A 1,4-benzodiazepinone with a 1,3-dihydro-2H-1,4-benzodiazepin-2-one core, this molecule features a 2-(diethylamino)ethyl group, 2-fluorophenyl group, and chloro group at positions 1, 5, and 7, respectively. As a partial agonist of GABAA receptors, it exhibits sedative, anticonvulsant, and anxiolytic effects. Its therapeutic applications include the treatment of insomnia, making it a valuable pharmaceutical agent in managing sleep disorders. & \small The molecule is a 1,4-benzodiazepinone derivative with a 2-(diethylamino)ethyl side chain and substitutions of a 2-fluorophenyl group at position 5 and a chloro group at position 7. Known for its partial agonist activity on GABAA receptors, it is commonly used as a sedative, anticonvulsant, and anxiolytic agent for the treatment of insomnia. Its chemical structure includes a diazepine ring fused to a benzene ring with specific substituents, making it a potent therapeutic compound with diverse pharmacological properties. \\ 
    \midrule
    \raisebox{-\totalheight}{\includegraphics[width=2.5cm,height=2.5cm]{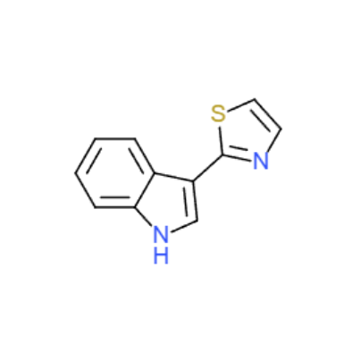}} & \small The molecule is an indole phytoalexin that is indole substituted at position 3 by a 1,3-thiazol-2-yl group. It has a role as a metabolite. It is an indole phytoalexin and a member of 1,3-thiazoles. & \small Indole phytoalexin with a 1,3-thiazol-2-yl substitution at position 3. It plays a role as a metabolite and exhibits indole phytoalexin and 1,3-thiazole properties. & \small The molecule is a biologically active compound containing an indole ring substituted at position 3 with a 1,3-thiazol-2-yl group. It acts as an indole phytoalexin and plays a role as a metabolite. Its structure consists of fused aromatic rings and a sulfur-containing heterocycle, making it a unique and potentially potent agent in biological systems. \\ 
    \bottomrule
\end{tabular}
}
\caption{
Example of molecules and their captions generated by GPT 3.5-turbo and Gemini Pro in \newdataset.
}
\label{table:rewritten_captions}
\end{table*}

This section presents some examples of augmented captions. 
Particularly, we provide the SMILES string, the original caption of \olddataset and four different captions rewritten by four LLMs, \ie, two open-source LLMs (Llama 2-70b~\cite{TMSA23} and Llama 3-70b~\cite{TMSA23}) and two closed-source LLMs (GPT 3.5-turbo~\cite{AAAA23} and Gemini Pro~\cite{Gemini24}). 

\begin{enumerate}[leftmargin=*]\itemsep0em
    \item 
    \textbf{SMILES string:} [18FH] \\
    \textbf{Original caption:} The molecule is the radioactive isotope of fluorine with relative atomic mass 18.000938. The longest-lived fluorine radionuclide with half-life of 109.77 min. \\
    \textbf{GPT 3.5-turbo:} The molecule is the radioactive isotope of fluorine with relative atomic mass 18.000938 and atomic number 9. It is the longest-lived fluorine radionuclide with half-life of 109.77 min. It is used in positron emission tomography (PET) for imaging the brain, heart, and other organs. \\
    \textbf{Gemini Pro:} This molecule is a radioactive isotope of fluorine with a relative atomic mass of 18.000938. It is the longest-lived fluorine radionuclide with a half-life of 109.77 minutes. \\
    \textbf{Llama 2-70b:} This molecule is a radioactive isotope of fluorine, with a relative atomic mass of 18.000938. It is the longest-lived fluorine radionuclide, with a half-life of 109.77 minutes. It has a number of applications in medical imaging and research, including positron emission tomography (PET) scans and fluorodeoxyglucose (FDG) tracers. \\
    \textbf{Llama 3-70b:} This molecule is fluorine-18 a radioactive isotope of fluorine with a relative atomic mass of 18.000938 and a half-life of 109.77 minutes making it the longest-lived fluorine radionuclide. \\
    \item 
    \textbf{SMILES string:} CCN(CC)CCN1C(=O)CN=C\\(C2=C1C=CC(=C2)Cl)C3=CC=CC=C3F \\
    \textbf{Original caption:} The molecule is a 1,4-benzodiazepinone that is 1,3-dihydro-2H-1,4-benzodiazepin-2-one substituted by a 2-(diethylamino)ethyl group, 2-fluorophenyl group and chloro group at positions 1, 5 and 7, respectively. It is a partial agonist of GABAA receptors and used for the treatment of insomnia. It has a role as a sedati gve, an anticonvulsant, a GABAA receptor agonist and an anxiolytic drug. It is a 1,4-benzodiazepinone, an organochlorine compound, a member of monofluorobenzenes and a tertiary amino compound. \\
    \textbf{GPT 3.5-turbo:} A 1,4-benzodiazepinone with a 1,3-dihydro-2H-1,4-benzodiazepin-2-one core, this molecule features a 2-(diethylamino)ethyl group, 2-fluorophenyl group, and chloro group at positions 1, 5, and 7, respectively. As a partial agonist of GABAA receptors, it exhibits sedative, anticonvulsant, and anxiolytic effects. Its therapeutic applications include the treatment of insomnia, making it a valuable pharmaceutical agent in managing sleep disorders. \\
    \textbf{Gemini Pro:} The molecule is a 1,4-benzodiazepinone derivative with a 2-(diethylamino)ethyl side chain and substitutions of a 2-fluorophenyl group at position 5 and a chloro group at position 7. Known for its partial agonist activity on GABAA receptors, it is commonly used as a sedative, anticonvulsant, and anxiolytic agent for the treatment of insomnia. Its chemical structure includes a diazepine ring fused to a benzene ring with specific substituents, making it a potent therapeutic compound with diverse pharmacological properties. \\
    \textbf{Llama 2-70b:} This molecule is a 1,4-benzodiazepinone, a type of sedative drug that selectively activates GABAA receptors. It has a 2-(diethylamino)ethyl group, 2-fluorophenyl group, and chloro group attached at positions 1, 5, and 7, respectively. Its role as a partial agonist of GABAA receptors makes it useful for treating insomnia, and it also has anticonvulsant, anxiolytic, and sedative properties. This compound belongs to the class of organochlorine compounds and is a tertiary amino compound. \\
    \textbf{Llama 3-70b:} This molecule is a 1,4-benzodiazepinone featuring a 2-(diethylamino)ethyl substituent at position 1, a 2-fluorophenyl group at position 5, and a chloro group at position 7, functioning as a partial GABAA receptor agonist used to treat insomnia, and possessing sedative, anticonvulsant, and anxiolytic properties. \\
    \item
    \textbf{SMILES string:} C1=CC=C2C(=C1)C(=CN2)\\C3=NC=CS3 \\
    \textbf{Original caption:} The molecule is an indole phytoalexin that is indole substituted at position 3 by a 1,3-thiazol-2-yl group. It has a role as a metabolite. It is an indole phytoalexin and a member of 1,3-thiazoles. \\
    \textbf{GPT 3.5-turbo:} Indole phytoalexin with a 1,3-thiazol-2-yl substitution at position 3. It plays a role as a metabolite and exhibits indole phytoalexin and 1,3-thiazole properties. \\
    \textbf{Gemini Pro:} The molecule is a biologically active compound containing an indole ring substituted at position 3 with a 1,3-thiazol-2-yl group. It acts as an indole phytoalexin and plays a role as a metabolite. Its structure consists of fused aromatic rings and a sulfur-containing heterocycle, making it a unique and potentially potent agent in biological systems. \\
    \textbf{Llama 2-70b:} This indole phytoalexin features a unique molecular structure where an indole ring is substituted at position 3 by a 1,3-thiazol-2-yl group, resulting in a distinct chemical compound with significant biological activity. As a metabolite, it plays a crucial role in various physiological processes and is also a member of the 1,3-thiazole class of compounds, which are characterized by their sulfur-containing heterocyclic ring. \\
    \textbf{Llama 3-70b:} This molecule is a substituted indole featuring a 1,3-thiazol-2-yl group at position 3, classified as an indole phytoalexin and a metabolite, also belonging to the class of 1,3-thiazoles. \\
    \item 
    \textbf{SMILES string:} CN(CC(=O)O)N=O \\
    \textbf{Original caption:} The molecule is a nitrosamine that is sarcosine in which the hydrogen attached to the nitrogen has been replaced by a nitroso group. It has a role as a carcinogenic agent. It is a non-proteinogenic amino acid derivative and a nitrosamine. It derives from a sarcosine. \\
    \textbf{GPT 3.5-turbo:} The molecule is a nitrosamine that is sarcosine in which the hydrogen attached to the nitrogen has been replaced by a nitroso group. It has a role as a carcinogenic agent. It is a non-proteinogenic amino acid derivative and a nitrosamine. It derives from a sarcosine.	This molecule is a nitrosamine containing a nitroso group attached to a sarcosine derivative. It is known for its carcinogenic properties and is classified as a non-proteinogenic amino acid derivative. \\
    \textbf{Gemini Pro:} Sarcosine N-nitrosamide, a nitrosamine with a nitroso group replacing the hydrogen atom on the nitrogen atom of the sarcosine molecule, exerts carcinogenic effects. This non-proteinogenic amino acid derivative exhibits a potential role as a carcinogen and is derived from sarcosine. \\
    \textbf{Llama 2-70b:} This nitrosamine is a modified form of sarcosine, where the hydrogen atom attached to the nitrogen has been replaced by a nitroso group, resulting in a distinct chemical structure. As a non-proteinogenic amino acid derivative, it exhibits unique properties and functions, and its presence has been linked to carcinogenic effects. \\
    \textbf{Llama 3-70b:} This molecule is a carcinogenic nitrosamine derived from sarcosine where a nitroso group replaces the hydrogen attached to the nitrogen atom making it a non-proteinogenic amino acid derivative. \\
    \item 
    \textbf{SMILES string:} CC(=O)N[C@@H]1[C@H]\\([C@@H]([C@H](O[C@H]1O)CO)O[C@@H]\\2[C@@H]([C@H]([C@@H]([C@H](O2)CO)\\O)O[C@@H]3[C@H]([C@H]([C@@H]([C@\\H](O3)CO)O)O)O)NC(=O)C)O \\
    \textbf{Original caption:} The molecule is an amino trisaccharide consisting of alpha-D-mannopyranose, 2-acetamido-2-deoxy-alpha-D-glucopyranose and 2-acetamido-2-deoxy-beta-D-glucopyranose residues joined in sequence by (1->3) and (1->4) glycosidic bonds. It is a glucosamine oligosaccharide, an amino trisaccharide and a member of acetamides. It derives from an alpha-D-Manp-(1->3)-alpha-D-GlcpNAc and an alpha-D-GlcpNAc-(1->4)-beta-D-GlcpNAc. \\
    \textbf{GPT 3.5-turbo:} The molecule is an amino trisaccharide consisting of alpha-D-mannopyranose, 2-acetamido-2-deoxy-alpha-D-glucopyranose and 2-acetamido-2-deoxy-beta-D-glucopyranose residues joined in sequence by (1->3) and (1->4) glycosidic bonds. It is a glucosamine oligosaccharide, an amino trisaccharide and a member of acetamides. It derives from an alpha-D-Manp-(1->3)-alpha-D-GlcpNAc and an alpha-D-GlcpNAc-(1->4)-beta-D-GlcpNAc.	The molecule is a complex amino trisaccharide composed of alpha-D-mannopyranose, 2-acetamido-2-deoxy-alpha-D-glucopyranose, and 2-acetamido-2-deoxy-beta-D-glucopyranose residues linked together via (1->3) and (1->4) glycosidic bonds. This glucosamine oligosaccharide is a member of acetamides and plays a crucial role in various biological processes due to its unique sugar composition and glycosidic linkages. \\
    \textbf{Gemini Pro:} This molecule is an amino trisaccharide composed of an alpha-D-mannopyranose unit, a 2-acetamido-2-deoxy-alpha-D-glucopyranose unit, and a 2-acetamido-2-deoxy-beta-D-glucopyranose unit. These units are linked by (1->3) and (1->4) glycosidic bonds. This amino trisaccharide is a member of the acetamides family and is derived from an alpha-D-Manp-(1->3)-alpha-D-GlcpNAc and an alpha-D-GlcpNAc-(1->4)-beta-D-GlcpNAc. \\
    \textbf{Llama 2-70b:} This molecule is a glucosamine oligosaccharide composed of three residues: alpha-D-mannopyranose, 2-acetamido-2-deoxy-alpha-D-glucopyranose, and 2-acetamido-2-deoxy-beta-D-glucopyranose. These residues are joined in sequence by (1->3) and (1->4) glycosidic bonds. It is an amino trisaccharide, derived from an alpha-D-Manp-(1->3)-alpha-D-GlcpNAc and an alpha-D-GlcpNAc-(1->4)-beta-D-GlcpNAc. \\
    \textbf{Llama 3-70b:} This molecule is a trisaccharide composed of alpha-D-mannopyranose, 2-acetamido-2-deoxy-alpha-D-glucopyranose, and 2-acetamido-2-deoxy-beta-D-glucopyranose residues linked by (1->3) and (1->4) glycosidic bonds, classified as a glucosamine oligosaccharide and an acetamide derivative. \\
\end{enumerate}

\section{Evaluation Setup}
\label{sec:appendix_evaluation}

Following the benchmark settings of \olddataset, we train \newmodel on the training dataset of \newdataset and evaluate it on the test dataset. 
Since we are considering two molecular tasks: text-based \emph{de novo} molecule generation and molecule captioning, we employ two evaluation metric sets. 
 
To evaluate the molecule generation task, we employ eight metrics following previous work~\cite{ELRHCJ22}: SMILES \textbf{BLEU} score~\cite{PRWZ02}, \textbf{Levenshtein} distance~\cite{MVM09}, Fréchet ChemNet Distance (\textbf{FCD})~\cite{PRUHK18}, \textbf{MACCS FTS}~\cite{DLHN02}, \textbf{RDK FTS}~\cite{SSL15} \textbf{Morgan FTS}~\cite{RH10}, \textbf{Exact} score~\cite{ELRHCJ22}, and \textbf{Validity}~\cite{ELRHCJ22}.
Notably, there are three fingerprint metrics: MACCS FTS, RDK FTS and Morgan FTS. 
FTS stands for fingerprint Tanimoto similarity. 
MACCS, RDK, and Morgan are each fingerprinting methods for molecules. 
The fingerprints of two molecules are compared using Tanimoto similarity, and the average similarity over the evaluation matrix is reported. 
Additionally, we report exact SMILES string matches \ie, Levenshtein distance and SMILES BLEU score. 
Exact score and Validity are the percentage of generated molecules that exactly match the ground truth and the percentage of generated strings that are valid.

To evaluate the molecule captioning task, we employ three natural language generation metrics, \eg, Caption \textbf{BLEU} score~\cite{PRWZ02}, \textbf{ROUGE}~\cite{L04}, and \textbf{METEOR}~\cite{BL05}. 
BLEU measures the precision of n-grams between generated and reference texts, ROUGE evaluates recall and precision of overlapping units such as n-grams or word sequences, and METEOR combines precision, recall, and synonym matching for a more holistic evaluation of text generation quality.

Furthermore, the cross-modal evaluation metric \textbf{Text2Mol}~\cite{EZJ21} aims to train a retrieval model to rank molecules given their text descriptions. 
Unlike traditional metrics that rely on words or n-grams, the ranking function of Text2Mol uses cosine similarity between the ground truth molecule/description and the generated description/molecule, respectively. 
It can offer a more integrated assessment to measure the comprehensive semantics of molecule/description. 
Therefore, we adopt this metric as an essential assessment to understand the effectiveness of different models. 

\section{Baseline Models Description}
\label{sec:appendix_baseline}

This section presents brief descriptions of baseline models included in this work. 

\spara{RNN}~\cite{CMGBBSB14}.
It introduces a novel approach for improving statistical machine translation through the use of Recurrent Neural Networks (RNNs). 
They propose an encoder-decoder architecture that learns continuous-space representations for phrases. The encoder processes an input phrase and compresses it into a fixed-dimensional vector, while the decoder uses this vector to generate the target phrase. 
This method allows for better handling of variable-length input sequences and capturing long-term dependencies in phrases, leading to significant improvements in translation quality compared to traditional models.

\spara{Transformer}~\cite{VSPUJGKP17}.
It introduces the Transformer model, a novel neural network architecture designed for sequence transduction tasks, such as machine translation. 
The Transformer model relies entirely on attention mechanisms to capture dependencies between input and output without using recurrent or convolutional layers. 
This self-attention mechanism allows for greater parallelization and better handling of long-range dependencies compared to previous models. 
They demonstrate that the Transformer achieves state-of-the-art performance on translation tasks, significantly improving both training speed and translation quality. 

\spara{T5}~\cite{RSRLNMZLL20}.
It presents the Text-to-Text Transfer Transformer (T5), a model designed to unify various NLP tasks by converting all tasks into a text-to-text format. 
They explore the capabilities of transfer learning within this framework, demonstrating that the same model, training objective, hyperparameters, and architecture can be applied across a wide range of NLP tasks. 
By pre-training on a massive and diverse dataset and fine-tuning specific tasks, T5 achieves state-of-the-art performance on numerous benchmarks. 
Additionally, they propose the \olddataset dataset. 

\spara{\oldmodel}~\cite{ELRHCJ22}.
It explores the novel concept of bridging the gap between molecular representations and natural language descriptions. 
They propose a model, \oldmodel, that translates molecular structures into textual descriptions and vice versa. 
This interdisciplinary approach leverages advances in natural language processing and cheminformatics, using techniques such as neural networks to encode and decode information between these two domains. 
\oldmodel is the fundamental model that motivates our work. 

\spara{Text+Chem T5}~\cite{CGBWLM23}. 
Text+Chem T5 is a novel multi-task, multi-domain language model designed to bridge the gap between natural language and chemical language tasks. 
Built on the T5 architecture, it is specifically designed to handle tasks spanning both textual and chemical domains. 
This model can effectively translate between natural and chemical languages, enabling it to perform a variety of tasks such as chemical reaction prediction (forward and retrosynthesis), text-conditional de novo molecule generation, molecular captioning, and paragraph-to-action conversion for chemical procedures.

\spara{TGM-DLM}~\cite{GLWW24}. 
TGM-DLM employs a Transformer-based architecture with cross-attention to incorporate textual guidance. 
It is trained using two objectives: denoising embeddings with text guidance and recovering uncorrupted SMILES strings from deliberately corrupted ones. 
This training strategy enhances the model's ability to generate valid and relevant molecular structures.
The model demonstrates superior performance compared to autoregressive models like MolT5-Base, achieving this without additional data resources or pre-training. 

\spara{MolReGPT}~\cite{LLFWLTL24}. 
MolReGPT is a novel framework leveraging LLMs like GPT to advance molecule discovery through molecule-caption translation. 
Unlike traditional methods, which rely heavily on domain experts, computational resources, or domain-specific pre-training, MolReGPT uses ICL few-shot learning. 
This approach enables LLMs to perform molecule understanding and text-based molecule generation by retrieving and learning from similar molecules and their descriptions from a local database. 



\spara{BioT5}~\cite{PZZWGWXY23}.
It is a pre-training framework designed to enhance drug discovery by integrating molecules, proteins, and natural language. 
This framework addresses limitations in current models, such as generating invalid molecular SMILES, underutilising contextual information, and treating structured and unstructured knowledge equally. 
BioT5 utilises SELFIES~\cite{KHNFG20} for robust molecular representations and extracts relevant knowledge from the context surrounding bio-entities in unstructured biological literature. 

\spara{MolXPT}~\cite{LZXWXQZL23}. 
It is a unified language model that integrates text and molecular representations for enhanced molecular modelling. 
MolXPT leverages the success of Generative Pre-trained Transformers (GPT) by pre-training on SMILES sequences wrapped in relevant textual context. 
This involves detecting molecule names in text, replacing them with corresponding SMILES, and thus allowing mutual information exchange between text and molecule representations. 

\spara{BioT5+}~\cite{PWGLFZXQY24}.
BioT5+ is designed to bridge the gap between molecular data and textual descriptions in biological research. 
Building upon the BioT5 framework, BioT5+ introduces several innovations, including the integration of IUPAC nomenclature for molecules, which enhances its ability to understand molecular structures in both scientific literature and formal representations like SMILES and SELFIES. 
By employing multi-task instruction tuning, BioT5+ can generalise across diverse biological tasks, such as classification, regression, and generation, making it versatile for applications ranging from molecule property prediction to drug discovery. 

\section{Additional Experimental Results and Discussion}
\label{sec:appendix_additional_results_and_discussion}

\spara{Anecdotal molecule generation examples.}
Table~\ref{table:example_generated_molecule} shows some example molecules generated by \oldmodel-Small and \newmodel-Small, and the ground-truth molecules from \newdataset. 
From these examples, we can find that \newmodel-Small can generate accurate molecules similar to the ground truth, while \oldmodel-Small is making mistakes. 
(1) is an interesting case since (\emph{i}) its ground truth SMILES string has a long length, $88$ characters for which \newmodel-Small is able to generate an exact match; (\emph{ii}) indicates that \newmodel-Small can understand crystalline solids, like indolylmethylglucosinolate, in the annotation. 
In another interesting case, (2), \newmodel-Small not only understands chemical compounds but also can understand chemical treatments, \eg, replacement, mentioned in the annotation. 
These examples showcase the superiority of \newmodel for the text-based \emph{de novo} molecule generation task.

\spara{Anecdotal molecule captioning examples.}
Table~\ref{table:example_generated_caption} shows examples of molecule descriptions generated by \oldmodel-Small and \newmodel-Small, alongside the ground-truth descriptions. \newmodel-Small can generate more accurate and detailed descriptions that align closely with the ground truth, whereas \oldmodel-Small often misses important semantic details.
%
In (1), \oldmodel-Small omits critical details about the specific role and structure. 
In contrast, \newmodel-Small correctly identifies the molecule as derived from malonic acid and mentions its role as an Escherichia coli metabolite and mouse metabolite.
These examples show the superiority of \newmodel-Small in generating detailed and accurate molecule descriptions, making it a more effective model for the molecule captioning task.

\section{Additional Experiments}
\label{sec:appendix_additional_experiments}

Section~\ref{subsec:preliminary} and Section~\ref{subsec:language_augmentation} demonstrate a detailed example implementation of \pipeline on the \olddataset for challenging molecular generation tasks. 
To further demonstrate the versatility of \pipeline, we extend its application to several additional datasets, \bace~\cite{HFZDRLCL20}, \hiv~\cite{HFZDRLCL20}, \esol~\cite{HFZDRLCL20}, and \image~\cite{SDGS18}. 
These datasets support a variety of crucial tasks, such as \emph{image captioning}, \emph{text understanding}, and \emph{graph property prediction}.

\subsection{Dataset and Task}
\label{subsec:appendix_dataset_and_task}

We consider four benchmark datasets, which contain \emph{image}, \emph{text} and \emph{graph} data. 
\begin{enumerate}[leftmargin=*]\itemsep0em
    \item \bace. The \bace dataset provides quantitative ($\mathrm{IC}_{50}$) and qualitative (binary label) binding results for a set of inhibitors of human b-secretase 1 (BACE-1). 
    All data are experimental values reported in the scientific literature over the past decade, some with detailed crystal structures available. 
    \textbf{Task:} \bace merged a collection of 1,522 compounds with their 2D structures and binary labels, built as a classification task. 
    \item \hiv. The HIV dataset was introduced by the Drug Therapeutics Program (DTP) AIDS Antiviral Screen, which tested the ability to inhibit HIV replication for 41,127 compounds. 
    Screening results were evaluated and placed into three categories: confirmed inactive (CI), confirmed active (CA) and confirmed moderately active (CM).
    We further combine the latter two labels, making it a classification task between inactive (CI) and active (CA and CM). 
    \textbf{Task:} As we are more interested in discovering new categories of HIV inhibitors based on the available \emph{text} and \emph{graph} structure information.  
    \item \esol. \esol is a small dataset consisting of water solubility data for 1,128 compounds. 
    \textbf{Task:} We intend to estimate solubility directly from chemical \emph{graph} structures (as encoded in \emph{text} SMILES strings). 
    \item \image. \image is a large-scale dataset comprising around $3.3$ million image-caption pairs. 
    It is designed for automatic image captioning tasks and represents a significant step forward in terms of the variety and volume of data compared to previous datasets like MS-COCO. 
    \textbf{Task:} We follow the settings of \cite{FKIKT23} to train CLIP model~\cite{RKHRGASAMCKS21} and test it on ImageNet~\cite{DDSLLF09}. 
\end{enumerate}

\subsection{Automatic Annotation Augmentation}
\label{subsec:appendix_auto_annotation_augmentation}

Given \bace, \hiv, and \esol datasets, we first generate descriptions following the instruction of \cite{ZZM24}. 
Consequently, we query LLMs to augment these descriptions as described in Section~\ref{subsec:language_augmentation}. 
The prompt is designed as follows:
\begin{mybox}
    \textbf{Instruction: }
    You are now a chemical specialist in rewriting descriptions for a molecule in SMILES format. 
    Make sure those descriptions describe the given molecule correctly and precisely based on your two inputs (SMILES and Description of it). 
    Also, make sure your rewriting captions do not include the input SMILES.
    %
    \\
    \textbf{Message: }
    \emph{SMILES string of target molecule:} [SMILES]. \\
    \emph{Description of the molecule:} [description]. \\
    \emph{Task:} Rewrite the following molecule with its SMILES and description. 
    The newly rewritten caption should be elaborate, descriptive, and concise, highlighting the key structural features and biological activities of the molecule. 
    Only output rewritten caption without any header and linebreak. \\
    \emph{Answer:}
\end{mybox}

\image has available annotations for each image. 
We leverage LLMs to augment their annotations using this prompt:
\begin{mybox}
    \textbf{Instruction: }
    You are now a specialist in rewriting descriptions for an image. 
    Make sure those descriptions describe the given image correctly and precisely. 
    \\
    \textbf{Message: }
    \emph{Description of the image:} [description]. \\
    \emph{Task:} Rewrite the following description. 
    The newly rewritten caption should be elaborate, descriptive, and concise, highlighting the key knowledge of the molecule. 
    Only output rewritten caption without any header and linebreak. \\
    \emph{Answer:}
\end{mybox}

In this paper, we utilise two closed-source LLMs (GPT 3.5-turbo~\cite{AAAA23} and Gemini Pro~\cite{Gemini24}) to generate two rewritten annotations for the above-mentioned datasets.

\subsection{Training on Augmented Dataset}
\label{subsec:appendix_trainng_on_augmented_dataset}

After obtaining the augmented datasets (\bace, \hiv, and \esol), we simply combine three annotations of each molecule as the input features and integrate them within the LM and GNN models. 
Other training implementations follow the instruction of \cite{ZZM24}. 
About the \image dataset, we follow the implementation of \cite{FKIKT23} to integrate the augmented annotations with the CLIP model and evaluate them. 

Results in Table~\ref{table:results_more_dataset} show that \pipeline significantly enhances performance across these diverse applications. 
This improvement highlights \pipeline’s potential to be a valuable tool in a wide range of AI tasks, offering substantial gains in accuracy and efficiency.


\end{document}